\documentclass[conference]{IEEEtran}
\IEEEoverridecommandlockouts
\usepackage{cite}
\usepackage{amsmath,amssymb,amsfonts}
\usepackage{algorithmic}
\usepackage{graphicx}
\usepackage{textcomp}
\usepackage{xcolor}
\def\BibTeX{{\rm B\kern-.05em{\sc i\kern-.025em b}\kern-.08em
    T\kern-.1667em\lower.7ex\hbox{E}\kern-.125emX}}

\usepackage{algorithm}  
\usepackage{algorithmic}
\usepackage[utf8]{inputenc}
\usepackage{soul}
\usepackage{booktabs}

\usepackage{amsmath}
\usepackage{amsthm}
\usepackage{amssymb}
\usepackage{graphicx}

\usepackage{color}
\usepackage{xcolor}
\usepackage{subfigure}
\usepackage{bm}

\usepackage{colortbl}
\definecolor{mygray}{gray}{.8}
\definecolor{mypink}{rgb}{.99,.91,.95}
\definecolor{mycyan}{cmyk}{.3,0,0,0}

\def\etal{{\em et al.\/}\, }

\def\ie{\textit{i.e.}}
\usepackage[colorlinks,linkcolor=black,citecolor=black]{hyperref}

\usepackage{algorithm}
\usepackage{algorithmic}

\graphicspath{{figures/}}
\usepackage{multirow}

\def\ch{\textcolor{blue}}
\def\ch{\textcolor{black}}

\def\red{\textcolor{black}}

\usepackage{newfloat}
\usepackage{listings}
\floatstyle{ruled}
\newfloat{listing}{tb}{lst}{}
\floatname{listing}{Listing}

\begin{document}

\title{Two-level Graph Network for Few-Shot Class-Incremental Learning}

\author{
	\IEEEauthorblockN{
		Hao Chen$^{1*}$, 
		Linyan Li$^{2*}$, 
		Fan Lyu$^{3}$, 
		Fuyuan Hu$^{1,4,5\dagger}$,
		Zhenping Xia$^{1}$
		and Fenglei Xu$^{1}$} 
	\IEEEauthorblockA{$^{1}$ Suzhou University of Science and Technology, $^{2}$ Suzhou Institute of Trade \& Commerce, $^{3}$ Tianjin University, 
		\\ $^{4}$Jiangsu Industrial Intelligent and Low-carbon Technology Engineering Center, 
		\\ $^{5}$Suzhou Key Laboratory of Intelligent Low-carbon Technology Application
	\\  \{haochen@post, fuyuanhu@mail, xzp@mail, xufl@mail\}.usts.edu.cn, lilinyan@szjm.edu.cn, fanlyu@tju.edu.cn}
}


\maketitle

\begin{abstract}
Few-shot class-incremental learning (FSCIL) aims to design machine learning algorithms that can continually learn new concepts from a few data points, 
without forgetting knowledge of old classes. 
The difficulty lies in that limited data from new classes not only lead to significant overfitting issues but also exacerbates the notorious catastrophic forgetting problems. 
However, existing FSCIL methods ignore the semantic relationships between sample-level and class-level.
In this paper, we designed a two-level graph network for FSCIL named Sample-level and Class-level Graph Neural Network (SCGN).
Specifically, a pseudo incremental learning paradigm is designed in SCGN, which synthesizes virtual few-shot tasks as new tasks to optimize SCGN model parameters in advance.
Sample-level graph network uses the relationship of a few samples to aggregate similar samples and obtains refined class-level features.
Class-level graph network aims to mitigate the semantic conflict between prototype features of new classes and old classes.
SCGN builds two-level graph networks to guarantee the latent semantic of each few-shot class can be effectively represented in FSCIL.
Experiments on three popular benchmark datasets show that our method significantly outperforms the baselines and sets new state-of-the-art results with remarkable advantages.
Code is available at https://github.com/sukechenhao/SCGN.
\end{abstract}


\renewcommand{\thefootnote}{\fnsymbol{footnote}}
\footnotetext[1]{Co-first author.}
\footnotetext[2]{Corresponding author.}
\renewcommand{\thefootnote}{}

\section{Introduction}

In the real world, artificial intelligence often receives novel classes~\cite{zhou2021learning,du2022agcn}.
When updating the model with new classes, a fatal problem occurs, namely catastrophic forgetting~\cite{rebuffi2017icarl,lyu2021multi,sun2022exploring}, \ie, the discriminability of old classes drastically declines.
To meet the adaptation to new knowledge, Class-Incremental Learning (CIL)~\cite{pham2021dualnet,zhao2022deep,wang2022learning} recognizes new classes and maintains discriminability over old classes, which has become an important research area.
Most solutions to CIL problems are with abundant training samples.
However, in practical applications, the instance labeling and collection cost are sometimes unbearable, where the incremental class may have few samples. 
This CIL task with few training samples is called Few-Shot Class-Incremental Learning (FSCIL). 
Similar to CIL, learning new classes can lead to catastrophic forgetting of previous classes. 
In addition, due to the lack of new class instances, it is easy to observe the overfitting phenomenon on these limited inputs, 
which increases the learning difficulty of incremental tasks.

It is unwise to directly adopt CIL methods in FSCIL, where limited training samples result in serious overfitting and poor performance on old classes~\cite{tao2020few}.
In recent years, several works~\cite{zhang2021few,zhou2022forward} are designed for FSCIL, which classify FSL tasks through the class mean (prototype feature) to alleviate the problem of overfitting. 
However, these methods are difficult to distinguish few-shot classes well because only the prototype can hardly mine the latent semantic similarities and dissimilarities among a few samples.
In traditional FSL, Graph Neural Network (GNN) can express complex interactions between samples by performing feature aggregation from neighbors, and mining refined information from a few samples between support and query data. 
However, unlike traditional FSL, the training data of FSCIL is incremental and in sequence, and the data of past classes is unavailable.
That is, FSCIL not only needs to solve the few-shot problem, 
but also needs to overcome the semantic interference cross tasks. 
The GNN used in traditional FSL cannot effectively solve these two problems at the same time.

\begin{figure}[t]
	\centering
	\includegraphics[width=1.\linewidth]{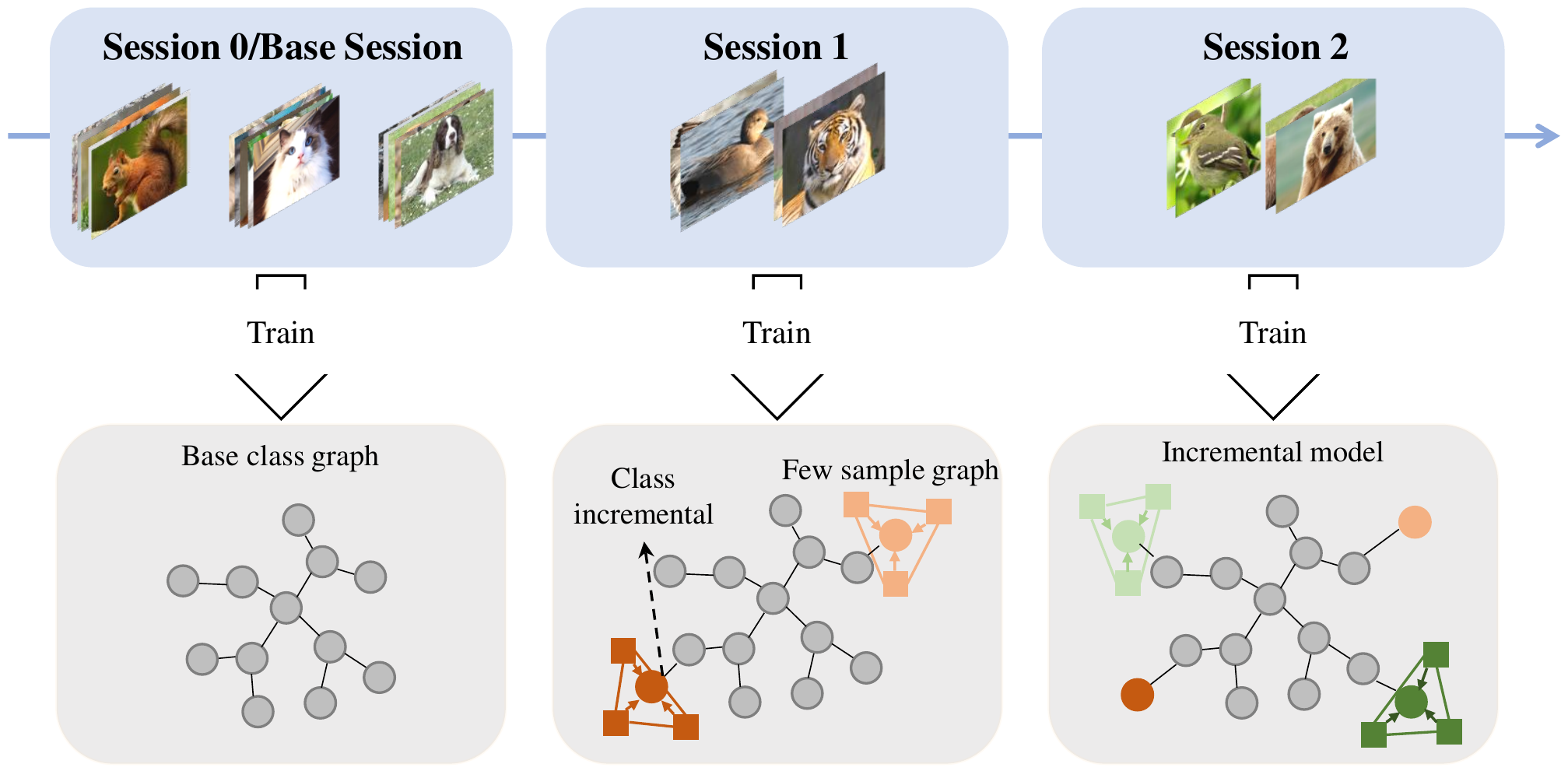}
	\caption{Illustration of our proposed two-level graph network for FSCIL. 
		Top: the setting of FSCIL.	
		Bottom: Sample-level to class-level graphs.
		\red{Square nodes represent sample-level features, and circular nodes represent class-level features.
		Sample-level features are learned through sample-level graph networks to obtain class-level features, 
		and class-level features are used to achieve class incremental learning through class-level graph networks.}
	}
	\label{fig:motivation}	
\end{figure}



In this paper, we propose a novel two-level graph network SCGN for FSCIL.
As shown in Fig.~\ref{fig:motivation}, the two levels are respectively Sample-level Graph network (SGN) and Class-level Graph Network (CGN).
\ch{
Specifically, we propose a pseudo incremental paradigm based on meta-learning to simulate FSCIL learning scenarios at the base training.
In the pseudo incremental process, we randomly sample FSL tasks from the base dataset, and generate virtual FSL tasks as new FSCIL tasks.
Then, in the process of meta-learning, SGN learns the FSL task, calculates the similarity between samples, gathers category samples, and distinguishes samples of different categories to obtain refined features.
Moreover, in order to alleviate the semantic gap between tasks, CGN calibrates the categories with the semantic gap according to the relationship 
between old classes and new classes to reduce the impact of new classes on old classes.
}
Experiments on benchmark datasets under various settings are conducted,
validating the effectiveness of our method.

\section{Related Work}

\textbf{Few-Shot Learning}.
Few-shot learning aims at rapidly generalizing to new tasks with limited samples, 
leveraging the prior knowledge learned from a large-scale base dataset.
The existing methods can be divided into two groups.
Optimization-based methods~\cite{lee2019meta,rusu2018meta} try to enable fast model adaptation with few-shot data.
Metric-based algorithms~\cite{zhang2020deepemd,ma2021transductive} utilize a pretrained backbone for feature extraction, and employ proper distance metrics between support and query instances.
Recent research tries to leverage GNNs to explore complex similarities among examples. 
DPGN~\cite{yang2020dpgn} builds up a dual graph to model distribution-level relations of examples for FSL. 
ECKPN~\cite{chen2021eckpn} proposes an end-to-end transductive GNN to explore the class-level knowledge.

\noindent
\textbf{Class-Incremental Learning}.
Class-Incremental Learning aims to learn from a sequence of new classes without forgetting old ones, which is now widely discussed in various computer vision tasks.
Current CIL algorithms can be divided into three groups.
The first group estimates the importance of each parameter and prevents important ones from being changed~\cite{aljundi2018memory,kirkpatrick2017overcoming}.
The second group utilizes knowledge distillation to maintain the model's discriminability~\cite{rebuffi2017icarl}.
Other methods rehearse former instances to overcome forgetting~\cite{zhao2020maintaining,zhu2021prototype}.
Pernici \etal~\cite{pernici2021class} pre-allocates classifiers for future classes, which needs extra memory for feature tuning and is unsuitable for FSCIL.

\noindent
\textbf{Few-Shot Class-Incremental Learning}
Few-Shot Class-Incremental Learning is recently proposed to address the few-shot inputs in the incremental learning scenario.
TOPIC~\cite{tao2020few} uses the neural gas structure to preserve the topology of features between old and new classes to resist forgetting.
\cite{cheraghian2021semantic} treats the word embedding as auxiliary information, and builds knowledge distillation terms to resist forgetting.
CEC~\cite{zhang2021few} utilizes an extra graph model to propagate context information between classifiers for adaptation.
FACT~\cite{zhou2022forward} efficiently incorporates new classes with forward compatibility and meanwhile resists the forgetting of old ones.

\begin{figure*}[t]
	\centering
	\includegraphics[width=1.\linewidth]{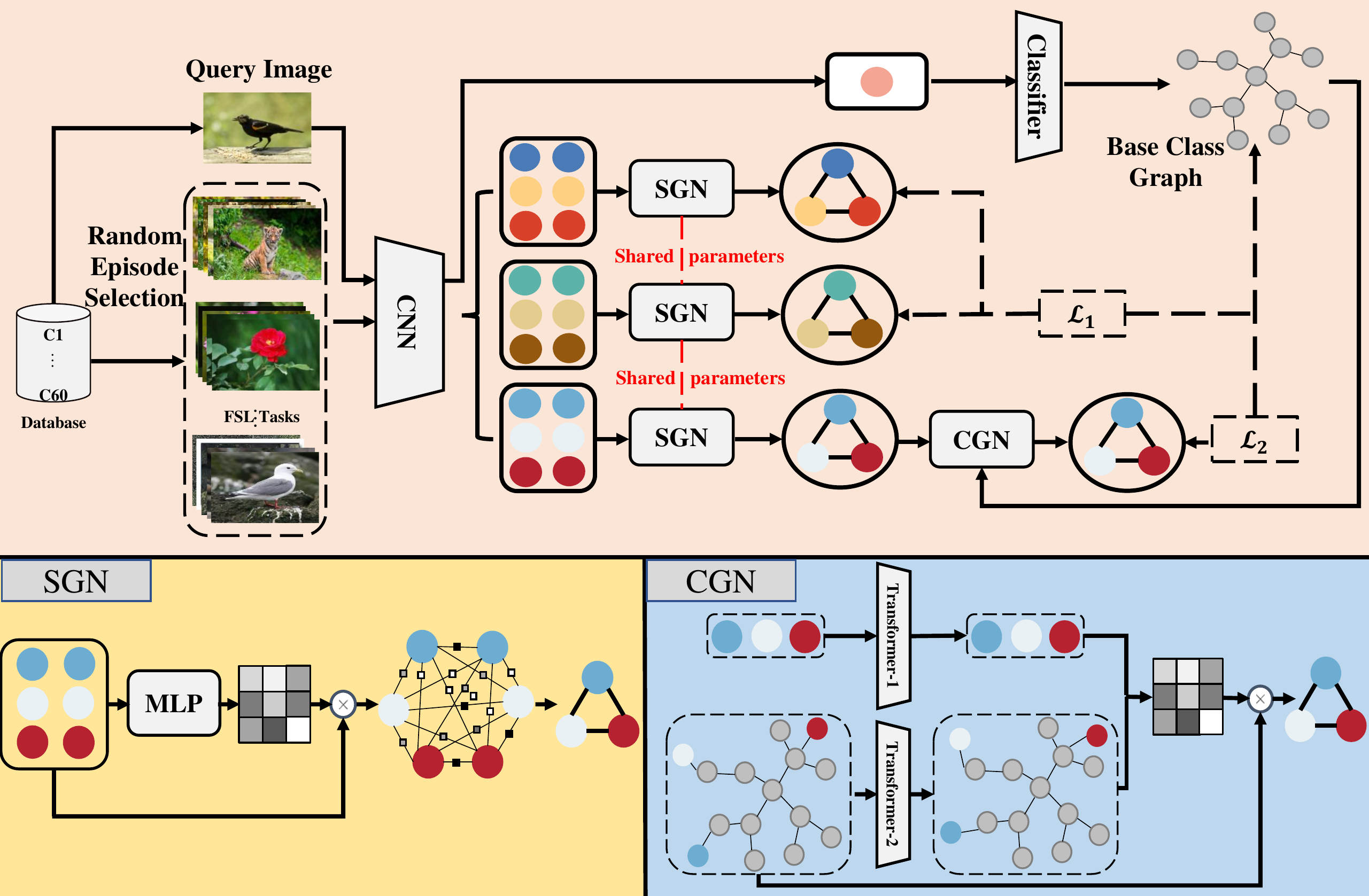}
	\caption{Our incremental prototype learning scheme for few-shot class-incremental learning.
		(a) an overview of the SCGN framework
		(b) Sample-Level Graph Network(SGN)
		(c) Class-Level Graph Network(CGN)}
	\label{fig:framework}
\end{figure*}

\section{Method}


\subsection{Problem Description and Pretraining}
\noindent
\textbf{Problem Description}.
We denote $ X $, $ Y $ and $ Z $ as the training set, the label set and the test set, respectively.
FSCIL task is to train a model from a continuous data stream in a class-incremental form, \ie, training sets  $ X^{0}, X^{1}, \dots X^{n} $, where samples of a set $ X^{i} $ are from the label set $ Y^{i} $,
and $ n $ represents the incremental session.
The incremental classes are disjoint, \ie, $ Y^{i} \bigcap Y^{j} = \varnothing $ for $i \neq j$.
For the base session, $ X^{0} $ has sufficient samples.
For each class in the subsequent sessions, we have only a few samples (e.g., 5 samples).
To measure a FSCIL model, we calculate the classification accuracy on the test set $ Z^{i} $ at each session $ i $.

\noindent
\textbf{Pretraining}. 
First, we will pretrain on the base dataset to obtain the class-level feature graph of the base session.
\red{In this case, the input of the model is only the query image $ Q $ to be predicted.}
We train a feature extractor $ f_{e} $ parameterized by $ \theta_{e} $ with a fully-connected layer as the classifier 
by minimizing the standard cross-entropy loss using the training samples of $ X^{0} $ under the supervision of target label $ T $.
We measure the relationship between the representation and the learnable class-level features $ \theta_{p} $
for all classes $ d(f_{e}(Q),\theta_{p}) $, $ d(\cdot,\cdot) $ represents the similarity measure, and we use the cosine similarity.
The pre-training optimization function can be expressed as:
\begin{equation}
	\theta_{\ast} = arg  \min\limits_{\theta} L(d(f_{e}(Q),\theta_{p}),T).
\end{equation}
Here $ \theta $ include above $ \theta_{e} $ and $ \theta_{p} $, $ L $ represents cross-entropy loss function.
To boost the ability of learning new classes in future tasks, we design a pseudo-incremental training paradigm at the base training based on meta-learning, which make the model learn how to learn a new class from a few samples.

\subsection{Two-level Graph Network (SCGN)}
\noindent
\textbf{Pseudo incremental learning}.
In the FSCIL task, the model should have the ability to adapt to new classes of knowledge and expand to new knowledge.
However, it is difficult to have the ability with only a few samples.
Therefore, we simulated the FSCIL learning situation and designed a pseudo-incremental learning paradigm in the base session to enhance the model's ability to adapt to new FSL tasks.
Specifically, we randomly sample two N-way K-shot (N classes, K samples for each class) FSL tasks, \ie, $ C_{1} $ and $ C_{2} $, from the base training set $ X^{0} $ in each iteration and we have $  Y^{c_{1}} \bigcap Y^{c_{2}} = \varnothing $.
These two FSL tasks serve as base tasks in the pseudo-incremental process. 

Motivated by~\cite{verma2019manifold}, we fuse instances by manifold mixup and treat the fused instances as virtual incremental classes.
We decouple the embedding into two parts at the hidden layer $ f_{e}(x) = g(h(x)) $.
We fuse two FSL tasks to generate a new virtual FSL task $ C_{3} $ :
\begin{equation}
	r_{i}^{c_{3}} = \sum_{i}\nolimits^{NK}g[\lambda h(x_{i}^{c_{1}}) + (1 - \lambda)h(x_{i}^{c_{2}})],
\end{equation}
where $ \lambda \in [0,1] $ is sampled from Beta distribution, 
$ r_{i}^{c_{3}} $ represents the features of the sample in the virtual FSL task.
The pseudo-incremental learning paradigm needs to enable two-level graph network to build graph relationships among samples and classes in FSCIL. 

\noindent
\textbf{Sample-Level Graph Network (SGN)}.
As shown in Fig.~\ref{fig:framework}, we obtained class-level graph of base task through pretraining.
Then, we introduce the Sample-level Graph Network (SGN) to learn the FSL task.
SGN aggregates samples of the same class and distinguishes samples of different classes by exploring the relationship between a few samples, so as to mine refined class-level features.
This not only improves the performance of FSL tasks, but also increases the extensibility of feature representations.
The formula for the relationship between samples in each FSL task is as follows:
\begin{equation}
	e_{ij}^{c} = f_{r}( (r_{i}^{c} - r_{j}^{c} )^2 ),
\end{equation}
where $ r_{i}^{c}, r_{j}^{c} $ respectively represents the sample features of the $i$-th and $j$-th FSL task $ C $,
$  C \in \left \{ C_{1},C_{2} \right \} $ and 
$ f_{r} $ is the encoding network that transforms the instance similarity to a certain scale.
$ f_{r} $ contains two Conv-BN-ReLU blocks. 
We update the sample representation through the relationship parameters between samples.
The obtained embeddings are averaged for each class as a class-level feature:
\begin{equation}
	R_{s}^{c} = \text{mean}(\text{SGN}(r_{i}^{c} + \sum_{j}\nolimits^{NK} e_{ij}^{c} \cdot r_{j}^{c} )) ,
\end{equation}
where $ \text{SGN} $ is the aggregation network with parameter set $ \theta_{s} $.




\noindent
\textbf{Class-Level Graph Network (CGN)}.
In the process of FSCIL incremental learning, the model should adjust itself with the new FSL task and perform well on old task.
However, the incremental model is optimized on the many-shot old classes, which is tailored to depict old classes' features.
As a result, there exists a semantic gap between the old classifiers and extracted new classes prototypes.
To solve the semantic gap between the old class and the new class,
we introduce the Class-level Graph Network (CGN).


\begin{figure*}[t]
	\centering
	\includegraphics[width=1\linewidth]{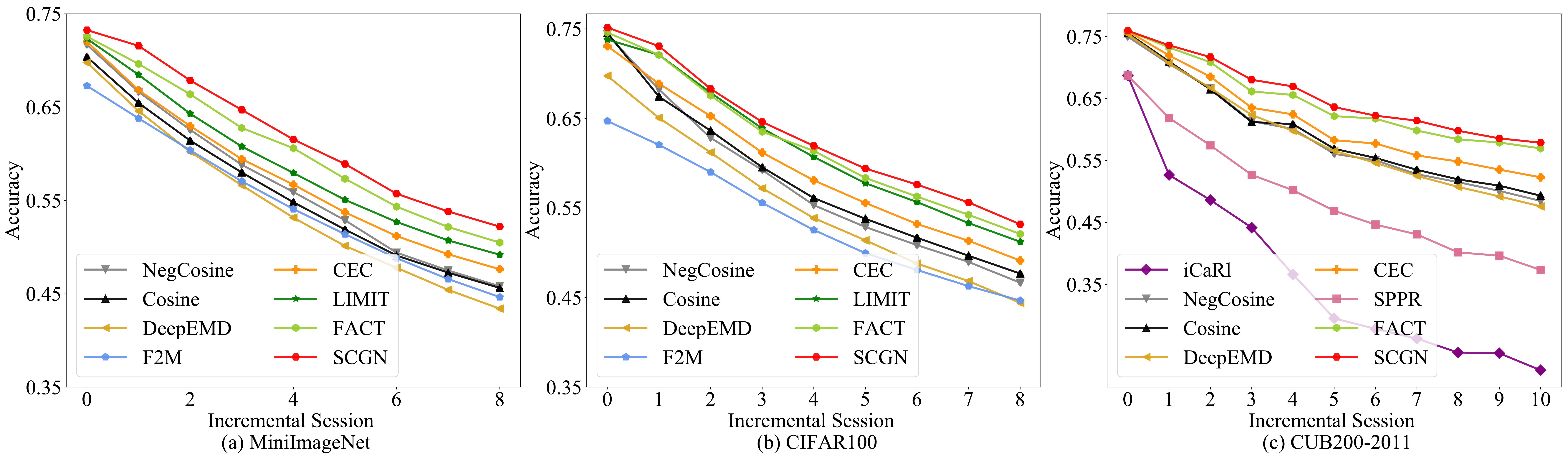}
	\caption{Comparison of our classification results with other methods on MiniImageNet, CIFAR100 and CUB200-2011.
	From the experimental results, it can be seen that SCGN outperforms the state-of-the-art (SOTA) methods.}
	\label{fig:comparison}
\end{figure*}

CGN should reflect the context relationship between old and new classes, 
so as to adjust the embedding space of prototype features of new classes in the class-level graph.
In our implementation, we combine the Transformer~\cite{vaswani2017attention} with the GNN.
Specifically, we use the multi-head attention mechanism to construct the relationship between the old class and the new class, 
and use the GNN to aggregate these information to calibrate the prototype features of the new class.
Transformer is a store of triplets in the form of (query $ Q $, key $ K $ and value $ V $). We set these parameters to
$ V =  [\theta_{p}, R_{s}^{c_{3}}] $,
$ K = W_{k}^{T} R_{s}^{c_{3}} $,
$ Q = W_{q}^{T} V $.
$ R_{s}^{c_{3}} $ is class-level features obtained by SGN learning virtual FSL task $ C_{3} $.
$ W_{k}$ and $ W_{q} $ are the learnable parameter of linear projection function.
The class-level features formula after CGN calibrating FSL $ C_{3} $ is as follows:
\begin{equation}
	\widetilde{R_{s}}^{c_{3}}  = \text{CGN} ( R_{s}^{c_{3}} + \sum\nolimits_{k} \alpha_{kq} \cdot V_{k} ) .
\end{equation}
where $ \alpha_{kq}  \propto \text{exp}( \frac{K Q^{T}}{\sqrt{d} } )  $ 
represents the association weight between the old class features and the new class features,
$ \text{CGN} $ is the aggregation network with parameter set $ \theta_{c} $.

\subsection{FSCIL Training using SCGN}
Fig.~\ref{fig:framework} shows the training schematic of SCGN.
SGN matches the class-level features after learning with the base class graph, 
which not only strengthens SGN's ability to learn FSL tasks but also reduces the interference to other classes. 
CGN extends the calibrated class-level features to the base class graph and predicts the virtual samples constructed. 
Ensure performance while mitigating interference to old classes.
We define the following loss function to learn SGN:
\begin{equation}
	\mathcal{L}_{1} = - \sum\nolimits_{i}^{2N}\text{cos}(\text{tanh}(R_{s}^{c_1} \cup R_{s}^{c_2}),\text{tanh}(\theta_{p})).
\end{equation}
With continuous optimization, sample features of the same class in the base dataset will become more compact.
To keep the distinction between the new class and the old class, we define the following loss function to learn CGN:
\begin{equation}
	\mathcal{L}_{2} = L(d( r_{i}^{c_{3}} ,[\theta_{p}, \widetilde{R_{s}}^{c_{3}} ]),T),
\end{equation}
where $ [\cdot] $ denotes the concatenation operation, $ L $ represents cross-entropy loss function,
$ T $ represents the label of the virtual sample constructed.

%
%

\section{Experiment}

\begin{table*}[h]
	\centering
	\caption{Comparison with the state-of-the-art on MiniImageNet dataset.}\label{lab:mini}
	\resizebox{1.\linewidth}{!}{
		\begin{tabular}{l cccccccccc cc  }
			\bottomrule
			\multirow{2}{*}{\textbf{Methods}} &\multicolumn{9}{c}{\textbf{Accuracy in each session(\%)}}  & \multirow{2}{*}{\textbf{PD}} & \multicolumn{1}{c}{\textbf{Our relative}} \\
			\cline{2-10}
			&\textbf{0} &\textbf{1}  &\textbf{2}&\textbf{3} &\textbf{4}&\textbf{5}  &\textbf{6}&\textbf{7}&\textbf{8} & &\textbf{improvement}\\ 
			\hline														
			Finetune           &  61.31  &  27.22  &  16.37  &   6.08  &   2.54  &   1.56  &   1.93  &   2.60  &   1.40  &  59.91  &  \textbf{+38.66}  \\
			iCaRL~\cite{rebuffi2017icarl}            &  61.31  &  46.32  &  42.94  &  37.63  &  30.49  &  24.00  &  20.89  &  18.80  &  17.21  &  44.10  &  \textbf{+22.85}  \\
			EEIL~\cite{castro2018end}              &  61.31  &  46.58  &  44.00  &  37.29  &  33.14  &  27.12  &  24.10  &  21.57  &  19.58  &  41.73  & \textbf{+20.48}   \\
			Rebalancing~\cite{hou2019learning}        &  61.31  &  47.80  &  39.31  &  31.91  &  25.68  &  21.35  &  18.67  &  17.24  &  14.17  &  47.14  &  \textbf{+25.89}  \\
			TOPIC~\cite{tao2020few}              &  61.31  &  50.09  &  45.17  &  41.16  &  37.48  &  35.52  &  32.19  &  29.46  &  24.42  &  36.89  &  \textbf{+15.64}  \\
			Decoupled-Cosine~\cite{vinyals2016matching}   &  70.37  &  65.45  &  61.41  &  58.00  &  54.81  &  51.89  &  49.10  &  47.27  &  45.63  &  24.74  &  \textbf{+3.49}  \\
			Decoupled-DeepEMD~\cite{zhang2020deepemd}  &  69.77  &  64.59  &  60.21  &  56.63  &  53.16  &  50.13  &  47.79  &  45.42  &  43.41  &  26.36  &  \textbf{+5.11}  \\
			F2M~\cite{shi2021overcoming}                &  67.28  &  63.80  &  60.38  &  57.06  &  54.08  &  51.39  &  48.82  &  46.58  &  44.65  &  22.63  &  \textbf{+1.38}  \\
			CEC~\cite{zhang2021few}                &  72.00  &  66.83  &  62.97  &  59.43  &  56.70  &  53.73  &  51.19  &  49.24  &  47.63  &  24.37  &  \textbf{+3.12}  \\
			LIMIT~\cite{zhou2022few}             &  72.32  &  68.47  &  64.30  &  60.78  &  57.95  &  55.07  &  52.70  &  50.72  &  49.19  &  23.13  &  \textbf{+1.85}  \\
			FACT~\cite{zhou2022forward}               &  72.56  &  69.63  &  66.38  &  62.77  &  60.60  &  57.33  &  54.34  &  52.16  &  50.49  &  22.07  &  \textbf{+0.82}  \\
			\hline
			SCGN       	   &  \textbf{73.25}  &  \textbf{71.57}  &  \textbf{67.46}  &  \textbf{64.01}  &  \textbf{61.04}  &  \textbf{58.41}  &  \textbf{55.62}  &  \textbf{53.62}  &  \textbf{52.00}  &  \textbf{21.25}  &    \\
			\toprule																
	\end{tabular}}	
\end{table*}

\subsection{Implementation Details}
\noindent
\textbf{Dateset}: We evaluate on MiniImageNet, CUB200-2011 and CIFAR100.
MiniImageNet is a subset of ImageNet with 100 classes. CUB200-2011 is a fine-grained image classification task with 200 classes.
CIFAR100 contains 60,000 images from 100 classes. 

\noindent
\textbf{Dateset Split}: For MiniImageNet and CIFAR100, 100 classes are divided into 60 base classes and 40 new classes. 
The new classes are formulated into eight 5-way 5-shot incremental tasks.
For CUB200, 200 classes are divided into 100 base classes and 100 incremental classes, and the new classes are formulated into ten 10-way 5-shot incremental tasks. 

\noindent
\textbf{Compared methods}: We compare to classical CIL methods iCaRL~\cite{rebuffi2017icarl}, EEIL~\cite{castro2018end},
and Rebalancing~\cite{hou2019learning}. Besides, we also compare to current
SOTA FSCIL algorithms: TOPIC~\cite{tao2020few}, SPPR~\cite{zhu2021self},
Decoupled-DeepEMD/Cosine/NegCosine~\cite{liu2020negative,vinyals2016matching,zhang2020deepemd}, CEC~\cite{zhang2021few}, LIMIT~\cite{zhou2022few}
and FACT~\cite{zhou2022forward}. 

\noindent
\textbf{Training details}: All methods are implemented with Pytorch. 
For CIFAR100, we use ResNet20, while for others we use ResNet18. We optimize with SGD+momentum, and the learning rate is set to 0.1 and decays with cosine annealing.

\noindent
\textbf{Evaluation Protocol}: We evaluate models after each session on the test set $ Z^{i} $ and report the Top 1 accuracy.
We also use a performance dropping rate(\textbf{PD}) that measures the absolute accuracy drops in the last session w.r.t.
the accuracy in the first session,\ie, $ \text{PD} =  \mathcal{A}_{0} - \mathcal{A}_{N} $, 
where $ \mathcal{A}_{0} $ is the classification accuracy in the base session 
and $ \mathcal{A}_{N} $ is the accuracy in the last session.

\begin{figure}[t]
	\centering
	\includegraphics[width=1.\linewidth]{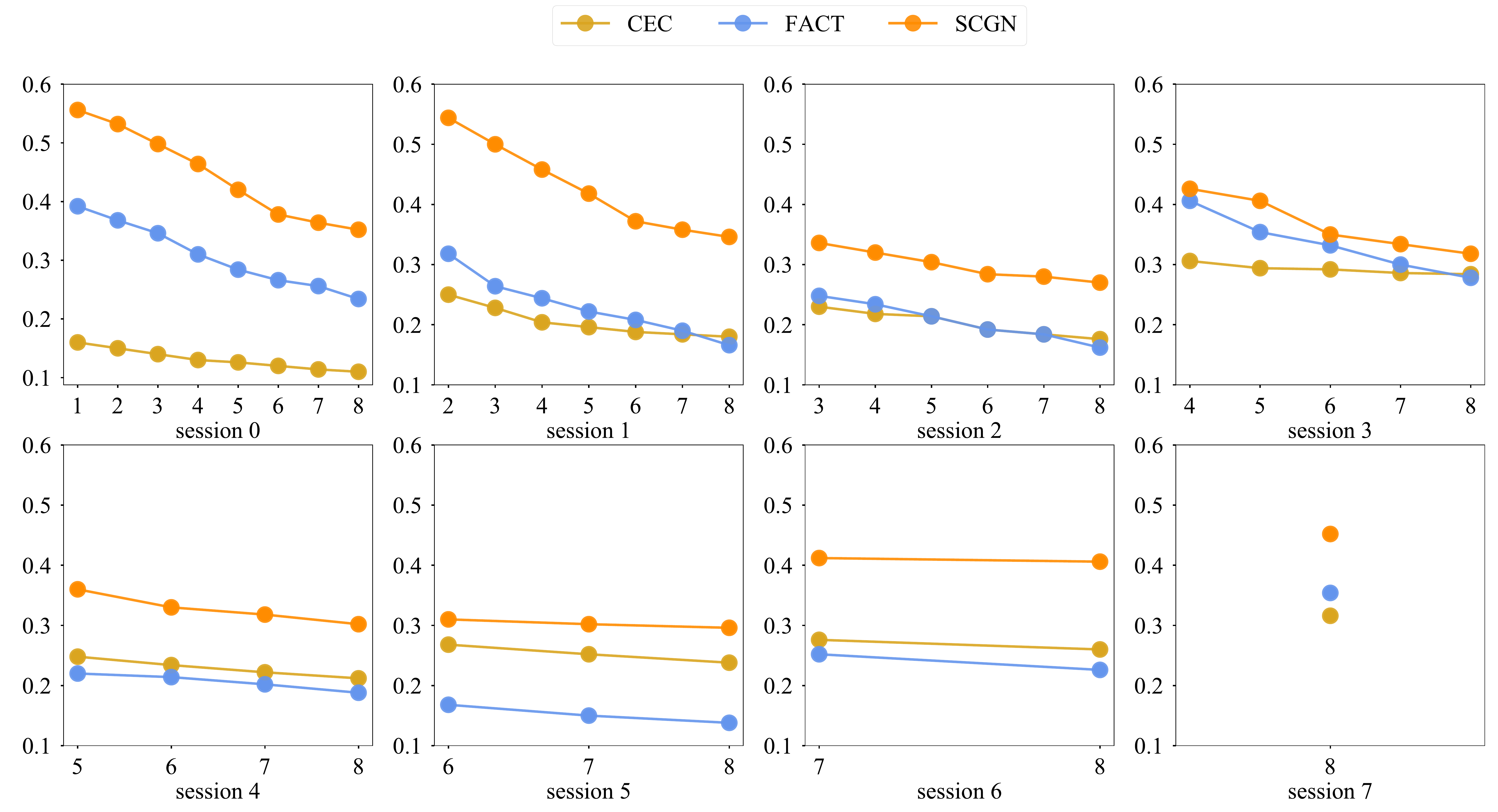}
	\caption{The accuracy of each session in MiniImageNet dataset is in the FSCIL task learning process.
	SCGN demonstrates superior performance in few-shot tasks within FSCIL tasks compared to other methods.}
	\label{fig:task_figures}
\end{figure}

\begin{figure}[t]
	\centering
	\includegraphics[width=1.\linewidth]{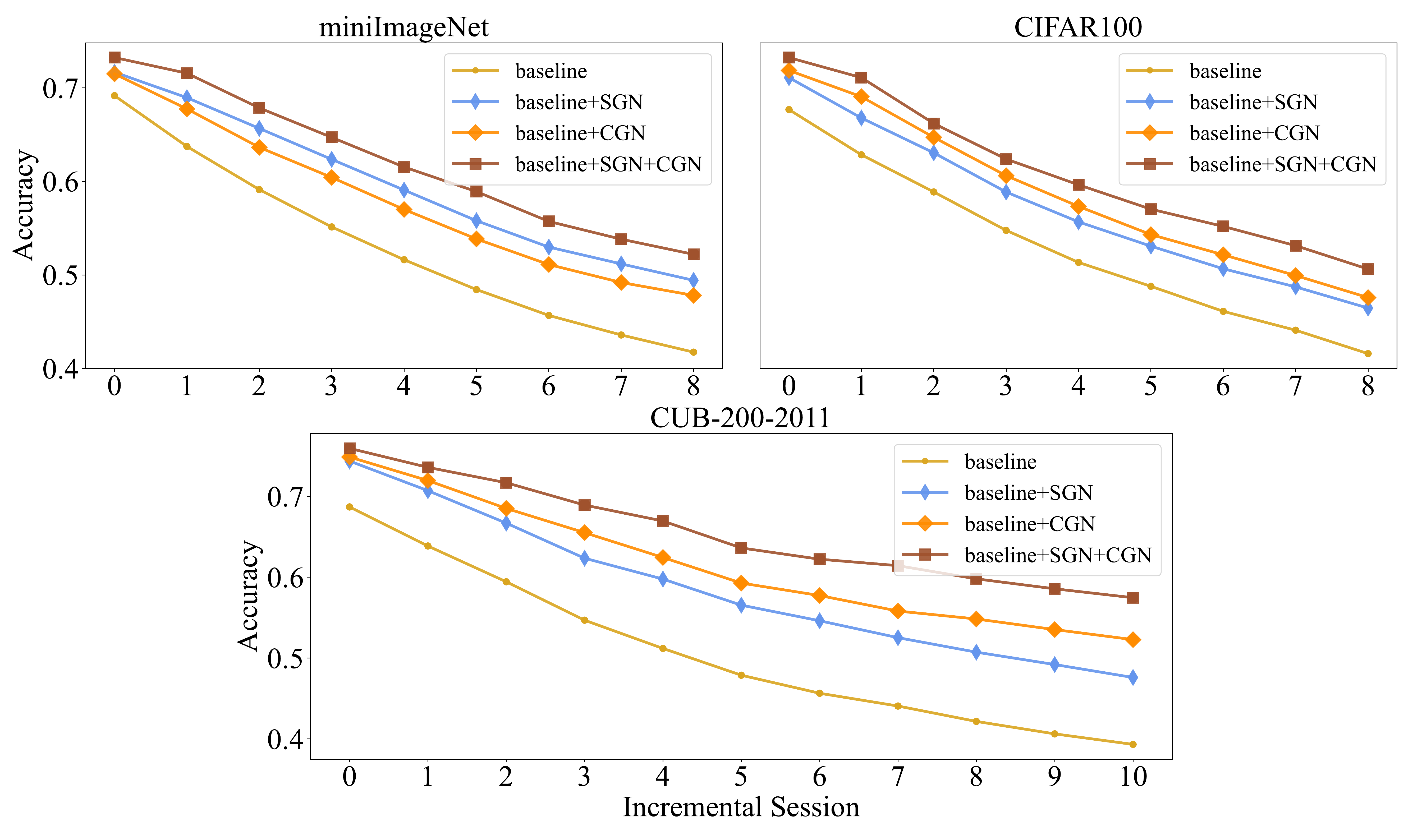}
	\caption{Ablation study on MiniImageNet, CIFAR100 and CUB-200-2011. Every part in SCGN improves the performance of FSCIL.}
	\label{fig:ablation}
\end{figure}

\subsection{Major Comparison}
We report the performance over benchmark datasets in Fig.~\ref{fig:comparison}.
We can infer from Fig.~\ref{fig:comparison} that SCGN consistently outperforms the current SOTA method, \ie, FACT~\cite{zhou2022forward} on benchmark datasets.
We also report the detailed value on MiniImageNet dataset in Table~\ref{lab:mini}. 
The performance of SCGN method is higher than that of other methods in each session, and the performance dropping rate is lower than that of other methods. 
The poor performance of CIL method (such as iCaRL) indicates that the method of a large number of sample tasks is not suitable for FSL tasks.
SCGN has better performance than Decoupled-DeepEMD/Cosine/NegCosine~\cite{liu2020negative,vinyals2016matching,zhang2020deepemd}, CEC~\cite{zhang2021few} and FACT~\cite{zhou2022forward}. 
It reveals that in FSCIL, it is important to make FSL tasks be trained well which strengthens new task constraints to reduce the impact on old tasks.
As shown in Fig.~\ref{fig:task_figures}, We compared the accuracy of each session on the MiniImageNet dataset with the CEC~\cite{zhang2021few} and FACT~\cite{zhou2022forward} methods.
It can be seen from the figure that in the FSCIL task learning process, the performance of each session is higher than that of other methods. 
This further proves the superiority of SCGN method.

\begin{figure}[t]
	\centering
	\includegraphics[width=1\linewidth]{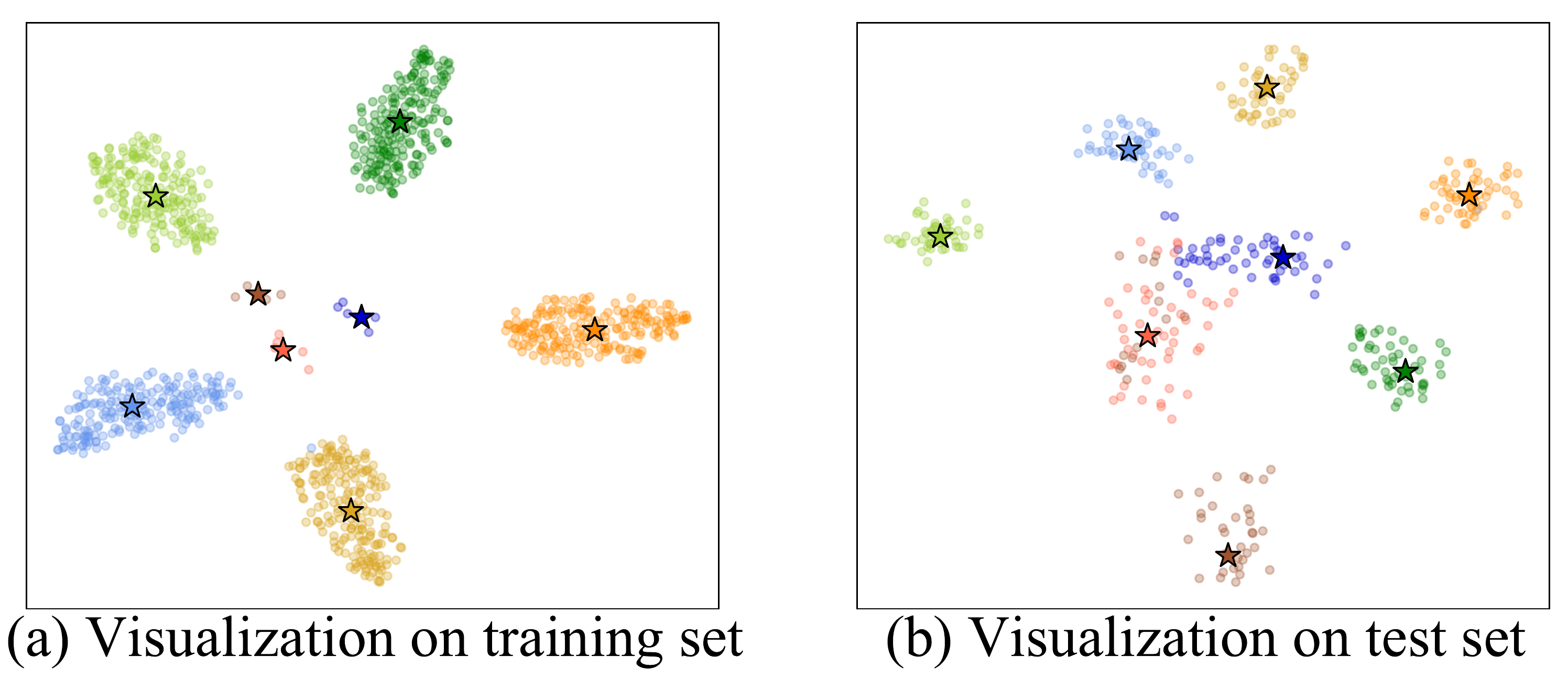}
	\caption{Visualization of decision boundary of training set and test set on CUB200-2011.
	\red{Circles represent sample features, stars represent class-level features, and different colors represent different categories.}}
	\label{fig:tsne}
\end{figure}

\subsection{Ablation Study}
We analyze the importance of each component of SCGN on MiniImageNet, CIFAR100 and CUB-200-2011 dataset in Fig.~\ref{fig:ablation}.
We separately construct models with different combinations of the core elements in SCGN.
Baseline represents that the backbone network is used to directly learn FSCIL tasks.
From Fig.~\ref{fig:ablation} we can infer that the use of CGN module effectively alleviates the catastrophic forgetting of incremental learning of baseline in FSCIL tasks.
The use of SGN module improves the learning performance of the FSL task, 
and significantly improves the overall performance of each session, which also proves the importance of the training of FSL tasks.
The combination of the two modules not only improves the learning performance of FSL tasks but also takes into account the semantic conflict between the old class and the new class due to data imbalance and other reasons. 
The ablation experiments validate that SGN and CGN modules are helpful for FSCIL tasks.

\subsection{Visualization of Incremental Session}
We visualize the learned decision boundaries with t-SNE on CUB-200-2011 dataset in Fig~\ref{fig:tsne}.
Fig.~\ref{fig:tsne}(a) stands for the decision boundary of the training set, where we train five old classes and three classes with few samples.
The circle represents the embedded space of the sample, and the star represents the class-level prototype.
We can find that a few samples of the new class are clustered, because the SGN learns more refined features through the association between samples.
In addition, CGN calibrates the categories with close similarity through the connection between the old class and the new class. 
It can be seen from the visualization that the class-level characteristics of the old class and the new class remain distinguishable
Fig.~\ref{fig:tsne}(b) tests the trained FSCIL task on the test set. 
It can be seen that SCGN helps to adapt the prototype and calibrate the decision boundary between old and new classes.

\section{Conclusion}
In this paper, we proposed a novel two-level graph network SCGN for FSCIL.
SCGN builds a pseudo incremental learning paradigm simulating FSCIL in base training.
SGN is used to build the relationship between samples in the FSL task to mine more favorable refined features, but also adapt to the learning paradigm of the FSCIL task, with strong model expansion capability.
CGN aligns cross tasks, solves the semantic gap between old classes and new classes, 
and alleviate the catastrophic forgetting problem in FSCIL tasks.
SCGN enhances the long-term learning ability of the model, making it consistent with the real scene.
Experimental results show that our model is superior in both performance and adaptability than the SOTA methods.


\bibliographystyle{IEEEtran}
	\bibliography{icme2023template}

\begin{thebibliography}{10}
\providecommand{\url}[1]{#1}
\csname url@samestyle\endcsname
\providecommand{\newblock}{\relax}
\providecommand{\bibinfo}[2]{#2}
\providecommand{\BIBentrySTDinterwordspacing}{\spaceskip=0pt\relax}
\providecommand{\BIBentryALTinterwordstretchfactor}{4}
\providecommand{\BIBentryALTinterwordspacing}{\spaceskip=\fontdimen2\font plus
\BIBentryALTinterwordstretchfactor\fontdimen3\font minus
  \fontdimen4\font\relax}
\providecommand{\BIBforeignlanguage}[2]{{%
\expandafter\ifx\csname l@#1\endcsname\relax
\typeout{** WARNING: IEEEtran.bst: No hyphenation pattern has been}%
\typeout{** loaded for the language `#1'. Using the pattern for}%
\typeout{** the default language instead.}%
\else
\language=\csname l@#1\endcsname
\fi
#2}}
\providecommand{\BIBdecl}{\relax}
\BIBdecl

\bibitem{zhou2021learning}
D.-W. Zhou, Y.~Yang, and D.-C. Zhan, ``Learning to classify with incremental
  new class,'' \emph{IEEE Transactions on Neural Networks and Learning Systems
  (TNNLS)}, 2021.

\bibitem{du2022agcn}
K.~Du, F.~Lyu, F.~Hu, L.~Li, W.~Feng, F.~Xu, and Q.~Fu, ``Agcn: augmented graph
  convolutional network for lifelong multi-label image recognition,'' in
  \emph{2022 IEEE International Conference on Multimedia and Expo
  (ICME)}.\hskip 1em plus 0.5em minus 0.4em\relax IEEE, 2022.

\bibitem{rebuffi2017icarl}
S.-A. Rebuffi, A.~Kolesnikov, G.~Sperl, and C.~H. Lampert, ``icarl: Incremental
  classifier and representation learning,'' in \emph{Proceedings of the
  IEEE/CVF Conference on Computer Vision and Pattern Recognition (CVPR)}, 2017.

\bibitem{lyu2021multi}
F.~Lyu, S.~Wang, W.~Feng, Z.~Ye, F.~Hu, and S.~Wang, ``Multi-domain multi-task
  rehearsal for lifelong learning,'' in \emph{Proceedings of the AAAI
  Conference on Artificial Intelligence (AAAI)}, vol.~35, no.~10, 2021.

\bibitem{sun2022exploring}
Q.~Sun, F.~Lyu, F.~Shang, W.~Feng, and L.~Wan, ``Exploring example influence in
  continual learning,'' 2022.

\bibitem{pham2021dualnet}
Q.~Pham, C.~Liu, and S.~Hoi, ``Dualnet: Continual learning, fast and slow,''
  \emph{Advances in Neural Information Processing Systems (NIPS)}, 2021.

\bibitem{zhao2022deep}
T.~Zhao, Z.~Wang, A.~Masoomi, and J.~Dy, ``Deep bayesian unsupervised lifelong
  learning,'' \emph{Neural Networks}, vol. 149, 2022.

\bibitem{wang2022learning}
Z.~Wang, Z.~Zhang, C.-Y. Lee, H.~Zhang, R.~Sun, X.~Ren, G.~Su, V.~Perot, J.~Dy,
  and T.~Pfister, ``Learning to prompt for continual learning,'' in
  \emph{Proceedings of the IEEE/CVF Conference on Computer Vision and Pattern
  Recognition (CVPR)}, 2022.

\bibitem{tao2020few}
X.~Tao, X.~Hong, X.~Chang, S.~Dong, X.~Wei, and Y.~Gong, ``Few-shot
  class-incremental learning,'' in \emph{Proceedings of the IEEE/CVF Conference
  on Computer Vision and Pattern Recognition (CVPR)}, 2020.

\bibitem{zhang2021few}
C.~Zhang, N.~Song, G.~Lin, Y.~Zheng, P.~Pan, and Y.~Xu, ``Few-shot incremental
  learning with continually evolved classifiers,'' in \emph{Proceedings of the
  IEEE/CVF Conference on Computer Vision and Pattern Recognition (CVPR)}, 2021.

\bibitem{zhou2022forward}
D.-W. Zhou, F.-Y. Wang, H.-J. Ye, L.~Ma, S.~Pu, and D.-C. Zhan, ``Forward
  compatible few-shot class-incremental learning,'' in \emph{Proceedings of the
  IEEE/CVF Conference on Computer Vision and Pattern Recognition (CVPR)}, 2022.

\bibitem{lee2019meta}
K.~Lee, S.~Maji, A.~Ravichandran, and S.~Soatto, ``Meta-learning with
  differentiable convex optimization,'' in \emph{Proceedings of the IEEE/CVF
  Conference on Computer Vision and Pattern Recognition (CVPR)}, 2019.

\bibitem{rusu2018meta}
A.~A. Rusu, D.~Rao, J.~Sygnowski, O.~Vinyals, R.~Pascanu, S.~Osindero, and
  R.~Hadsell, ``Meta-learning with latent embedding optimization,'' in
  \emph{International Conference on Learning Representations (ICLR)}, 2018.

\bibitem{zhang2020deepemd}
C.~Zhang, Y.~Cai, G.~Lin, and C.~Shen, ``Deepemd: Few-shot image classification
  with differentiable earth mover's distance and structured classifiers,'' in
  \emph{Proceedings of the IEEE/CVF Conference on Computer Vision and Pattern
  Recognition (CVPR)}, 2020.

\bibitem{ma2021transductive}
Y.~Ma, S.~Bai, S.~An, W.~Liu, A.~Liu, X.~Zhen, and X.~Liu, ``Transductive
  relation-propagation network for few-shot learning,'' in \emph{Proceedings of
  the Twenty-Ninth International Conference on International Joint Conferences
  on Artificial Intelligence (IJCAI)}, 2021.

\bibitem{yang2020dpgn}
L.~Yang, L.~Li, Z.~Zhang, X.~Zhou, E.~Zhou, and Y.~Liu, ``Dpgn: Distribution
  propagation graph network for few-shot learning,'' in \emph{Proceedings of
  the IEEE/CVF Conference on Computer Vision and Pattern Recognition (CVPR)},
  June 2020.

\bibitem{chen2021eckpn}
C.~Chen, X.~Yang, C.~Xu, X.~Huang, and Z.~Ma, ``Eckpn: Explicit class knowledge
  propagation network for transductive few-shot learning,'' in
  \emph{Proceedings of the IEEE/CVF Conference on Computer Vision and Pattern
  Recognition (CVPR)}, 2021.

\bibitem{aljundi2018memory}
R.~Aljundi, F.~Babiloni, M.~Elhoseiny, M.~Rohrbach, and T.~Tuytelaars, ``Memory
  aware synapses: Learning what (not) to forget,'' in \emph{Proceedings of the
  European conference on computer vision (ECCV)}, 2018.

\bibitem{zhao2020maintaining}
B.~Zhao, X.~Xiao, G.~Gan, B.~Zhang, and S.-T. Xia, ``Maintaining discrimination
  and fairness in class incremental learning,'' in \emph{Proceedings of the
  IEEE/CVF Conference on Computer Vision and Pattern Recognition (CVPR)}, 2020.

\bibitem{zhu2021prototype}
F.~Zhu, X.-Y. Zhang, C.~Wang, F.~Yin, and C.-L. Liu, ``Prototype augmentation
  and self-supervision for incremental learning,'' in \emph{Proceedings of the
  IEEE/CVF Conference on Computer Vision and Pattern Recognition (CVPR)}, 2021.

\bibitem{pernici2021class}
F.~Pernici, M.~Bruni, C.~Baecchi, F.~Turchini, and A.~Del~Bimbo,
  ``Class-incremental learning with pre-allocated fixed classifiers,'' in
  \emph{2020 25th International Conference on Pattern Recognition
  (ICPR)}.\hskip 1em plus 0.5em minus 0.4em\relax IEEE, 2021.

\bibitem{cheraghian2021semantic}
A.~Cheraghian, S.~Rahman, P.~Fang, S.~K. Roy, L.~Petersson, and M.~Harandi,
  ``Semantic-aware knowledge distillation for few-shot class-incremental
  learning,'' in \emph{Proceedings of the IEEE/CVF Conference on Computer
  Vision and Pattern Recognition (CVPR)}, 2021.

\bibitem{verma2019manifold}
V.~Verma, A.~Lamb, C.~Beckham, A.~Najafi, I.~Mitliagkas, D.~Lopez-Paz, and
  Y.~Bengio, ``Manifold mixup: Better representations by interpolating hidden
  states,'' in \emph{International conference on machine learning
  (ICML)}.\hskip 1em plus 0.5em minus 0.4em\relax PMLR, 2019.

\bibitem{vaswani2017attention}
A.~Vaswani, N.~Shazeer, N.~Parmar, J.~Uszkoreit, L.~Jones, A.~N. Gomez,
  {\L}.~Kaiser, and I.~Polosukhin, ``Attention is all you need,''
  \emph{Advances in Neural Information Processing Systems (NIPS)}, 2017.

\bibitem{castro2018end}
F.~M. Castro, M.~J. Mar{\'\i}n-Jim{\'e}nez, N.~Guil, C.~Schmid, and K.~Alahari,
  ``End-to-end incremental learning,'' in \emph{Proceedings of the European
  conference on computer vision (ECCV)}, 2018.

\bibitem{hou2019learning}
S.~Hou, X.~Pan, C.~C. Loy, Z.~Wang, and D.~Lin, ``Learning a unified classifier
  incrementally via rebalancing,'' in \emph{Proceedings of the IEEE/CVF
  Conference on Computer Vision and Pattern Recognition (CVPR)}, 2019.

\bibitem{vinyals2016matching}
O.~Vinyals, C.~Blundell, T.~Lillicrap, D.~Wierstra \emph{et~al.}, ``Matching
  networks for one shot learning,'' \emph{Advances in Neural Information
  Processing Systems (NIPS)}, 2016.

\bibitem{shi2021overcoming}
G.~Shi, J.~Chen, W.~Zhang, L.-M. Zhan, and X.-M. Wu, ``Overcoming catastrophic
  forgetting in incremental few-shot learning by finding flat minima,''
  \emph{Advances in Neural Information Processing Systems (NIPS)}, 2021.

\bibitem{zhou2022few}
D.-W. Zhou, H.-J. Ye, L.~Ma, D.~Xie, S.~Pu, and D.-C. Zhan, ``Few-shot
  class-incremental learning by sampling multi-phase tasks,'' \emph{IEEE
  Transactions on Pattern Analysis and Machine Intelligence (TPAMI)}, 2022.

\bibitem{zhu2021self}
K.~Zhu, Y.~Cao, W.~Zhai, J.~Cheng, and Z.-J. Zha, ``Self-promoted prototype
  refinement for few-shot class-incremental learning,'' in \emph{Proceedings of
  the IEEE/CVF Conference on Computer Vision and Pattern Recognition (CVPR)},
  2021.

\bibitem{liu2020negative}
B.~Liu, Y.~Cao, Y.~Lin, Q.~Li, Z.~Zhang, M.~Long, and H.~Hu, ``Negative margin
  matters: Understanding margin in few-shot classification,'' in
  \emph{Proceedings of the European conference on computer vision (ECCV)},
  2020.

\end{thebibliography}

\end{document}